\documentclass{article}

\usepackage[final]{nips_2016}

\usepackage[utf8]{inputenc} % allow utf-8 input
\usepackage[T1]{fontenc}    % use 8-bit T1 fonts
\usepackage{hyperref}       % hyperlinks
\usepackage{url}            % simple URL typesetting
\usepackage{booktabs}       % professional-quality tables
\usepackage{amsfonts}       % blackboard math symbols
\usepackage{nicefrac}       % compact symbols for 1/2, etc.
\usepackage{microtype}      % microtypography
\usepackage{amsmath}
\usepackage{graphicx}
\title{Positive blood culture detection in time series data using a BiLSTM network}

\author{
  Leen De Baets, Joeri Ruyssinck, Thomas Peiffer, \\ \textbf{Filip De Turck, Femke Ongenae, Tom Dhaene}  \\
  IBCN\\
  Ghent University - iMinds\\
  9052 Ghent, Belgium\\
  \texttt{leen.debaets@ugent.be} \\
  \And Johan Decruyenaere \\
  \\
  Ghent University Hospital \\
  Ghent University \\
  9000 Ghent, Belgium
}

\begin{document}

\maketitle

\begin{abstract}
The  presence  of  bacteria  or  fungi  in  the bloodstream  of  patients is abnormal and can  lead  to  life-threatening  conditions. A  computational model based on a bidirectional long short-term memory artificial  neural  network,  is  explored  to  assist  doctors  in  the intensive care unit to predict whether examination of blood  cultures  of  patients  will  return  positive. As  input  it  uses  nine monitored  clinical  parameters,  presented  as time series data, collected from $2177$ ICU admissions at the Ghent University Hospital. Our main goal is to determine if general machine learning methods and more specific, temporal models, can be used to create an early detection system. This preliminary research obtains an area of $71.95\%$ under the precision recall curve, proving the potential of temporal neural networks in this context.
\end{abstract}

\section{Introduction}
A positive blood culture is defined as a blood sample in which bacteria or fungi are present.  This growth of organisms in the blood stream can lead to inflammation
throughout  the  body  or  even  organ  failure  or  death [1].  When doctors suspect a patient to test positive they  can  decide  to advance to a blood culture test.  Symptoms indicative of a likely positive culture are complex and not fully understood.  Nevertheless, it is suspected a link exists between a patient's physiological data and the outcome of such a test.
 
Literature presents several techniques to detect sepsis [2,3,4,5] from patients physiological data. Sepsis is a condition related to a positive blood culture [6] and detection thereof could be similar to detecting positive blood cultures. Although the monitored patient data is time dependent, no models have been proposed in literature that specifically model the time aspect. This paper presents our work to explore the potential of temporal models to detect positive blood cultures.

\section{Data}

A database was constructed with physiological information from $2177$ patients admitted at the intensive care unit (ICU) of the Ghent University Hospital whereof $229$ admissions had a  positive  blood  culture  test.   For  all  other  patients, a blood test was performed which returned negative. For each patient, nine parameters were measured and calculated, these are listed in Table \ref{table:variables}. Each parameter is monitored with a different frequency. The total dataset contains more than fourteen million values.

First, we filter out outliers. We do this by defining bio-limit ranges for each variable (see Table \ref{table:variables}). Each value that falls outside this range is considered an outlier and removed. These outliers are caused by human error or machine malfunction and prove to be rare ($0.276\%$ of the data), as the database values are checked by study nurses. After removing the outliers, the data is normalised per variable using:
\begin{align}
n = \frac{x - \texttt{avg} }{3 * \texttt{std}}
\end{align}
where $x$ is the value, $ \texttt{avg}$ the average of all values and $ \texttt{std}$ is the standard deviation. 

As each of the variables in the database has its own monitoring frequency, this results in a different sequence length for each variable per patient. However, the method used in this paper (see Section \ref{sec:BiLSTM}) requires the sequence length of all variables to be equal. This is obtained by resampling the data. To define this, the total sequence time, sampling frequency, and sample end time need to be defined. We used the expertise of the medical experts involved in this research to initialise these parameters. Ideally, multiple settings and the effects of these parameters should be explored, but this lies beyond the scope of this initial study. More specifically, the total sequence time is configured to be $3$ days and the sampling frequency to one sample per hour. This results in a total of $72$ points per variable per patient. As end of the sampling period, we take  the moment when the first positive sample is established. If no positive sample is encountered , we choose as the sampling end-point the last available time point. The beginning of the sampled period is the end time minus $3$ days. If there is not enough data available for a patient (e.g.~if the admission only happened $2$ days before), then the data is padded with the means of the variables (zero because of the normalisation). If the sampling frequency of a variable is higher than one sample per hour, we subsample in such a way that the minimal, maximal or average value (depending on the variable, see Table \ref{table:variables}) is calculated in the sample window. If the sampling frequency is lower, we will repeat values.

In the end, there is a  time-sequence of $72$ points available for each patient where each point has $9$ features. A patient's label is one if it has a positive blood sample and the label is zero otherwise.

Recent research [7] handles the different monitoring frequencies by treating the formed gaps as features. As it lead to superior results in their case, future research should investigate this.

\begin{table}[t]
  \caption{Variables monitored per patient}
  \label{table:variables}
  \centering
  \begin{tabular}{lll}
    \toprule
    Variable     & bio-limits & sample approach\\
    \midrule
    Temperature & $[29 - 43]$  & max \\
    Blood thrombocyte count     &  & min      \\
    Blood leukocyte count     &        & mean \\
    C-Reactive Protein   &        & max\\
    Sepsis-related organ failure assessment  &        & max\\
    Heart rate  &      $[30 - 250]$     & max\\
    Respiratory Rare &       $[0 - 100]$     & max\\
    International Normalized Ratio of prothrombine time &       & max\\
    mean Systemic Arterial Pressure &   $[30 - 170]$     & max\\
    \bottomrule
  \end{tabular}
\end{table}

\section{Bidirectional LSTM} \label{sec:BiLSTM}
A  Recurrent  Neural  Network  (RNN)  is  a  computational model designed to work with temporal features. It is similar to a feed forward neural network with the extension that cycles are present in the network. Through those cycles, the network can implement memory, by allowing it to combine present inputs with inputs from several time steps in the past.

\begin{figure}[h]
  \centering
  \includegraphics[width = 0.7\linewidth]{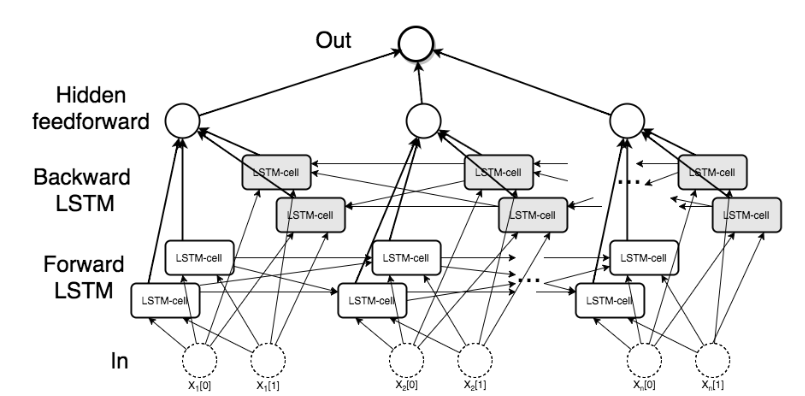} 
  \caption{Topology of an unfolded BiLSTM network with 2 imput features and 2 LSTM cells and an extra hidden feedforward layer.}
  \label{fig: BiLSTM}
\end{figure}

A  commonly  recognized  problem  in  training  recurrent  neural  networks  is  the  vanishing  gradient  problem.  The influence of inputs from several time steps fades away exponentially. This makes it impossible for those network to learn dependencies that span over long periods of time. Long Short-Time Memory (LSTM) networks [8] mitigate this problem by introducing the principle of gating. Conceptionally, these gates allow the network to implement small memory cell that is able to contain it's hidden state for longer periods of time, by blocking this cell's inputs and/or outputs. In a standard LSTM, information only flows in the forward time direction. A bidirectional LSTM (BiLSTM) also allows dependencies in the reverse direction, by combining two normal LSTMs, processing the sequence in both directions. Figure \ref{fig: BiLSTM} shows a schematic of a BiLSTM.

The basic network that is used for solving our problem has an input layer requesting the time sequence as a $72 \text{x} 9$ matrix. The input is then passed to one BiLSTM-layer that uses the $tanh$-function as activation function to introduce non-linearity. One single output is generated. This number is the prediction whether or not the given time sequence originates from a person with a positive blood culture or not. This is a floating point number, thus a threshold should be defined to binary classify the patient having a positive culture or not. We will not define a hard threshold. Rather, the precision recall curve is generated by varying this threshold.

To train the parameters of the network, the mean-squared error is used. Because the used data is imbalanced ($\#$ positives = $229$, $\#$ negatives = $1948$), the cost function is adapted in such a way that a larger error is given when a positive patient is wrongly classified compared to wrongly classifying a negative patient:
\begin{align}
\text{MSE} =  \sum_{i=1}^{n} w_{y,i} (\hat{y}_i - y_i)^2 
\end{align}
where $n$ are the amount of patients in the training set, $y_i$ is the label (positive or negative culture), $\hat{y}_i$ is the prediction, $ w_{y,i}$ is the class weight. This class weight is chosen such that patients with positive cultures are 8 times as important, since there are eight times as many patients with negative cultures.

\section{Results}
This section handles the evaluation of the network. Validation is done using the precision recall (PR) curve, which plots the precision against the recall. A good PR curve is defined by surface of the area it encloses, this is the so-called area-under-the-curve (AUC). The larger the AUC, the better. Compared to the  AUC of a receiving operating characteristic (ROC) curve, the AUC of the PR often provides a more clear metric of performance on imbalanced data.

For evaluation, the data is split into a training set ($90\%$) and a test set ($10\%$). This is done once in a stratified manner. On this training set, 10-fold cross validation is done to select the BiLSTM network with the optimal hyperparameters. The considered hyperparameters are the number of hidden nodes ($[10,100,1000]$), the learning rate ($[0.0001, 0.001, 0.01]$). The maximal number of epochs is $150$ but learning stops early if the PR AUC of the validation set is higher than $90 \%$ or when it lowers again. The optimal parameters are chosen such that the average of the PR AUC over the $10$ validation sets is maximal. The final model is an ensemble of the 10 models trained on the train data splits. Note, the division into different sets is done using stratified sampling guaranteeing that the proportion of positive samples in every set is equal.

\begin{figure}[h]
  \centering
  \includegraphics[width = 0.4\linewidth]{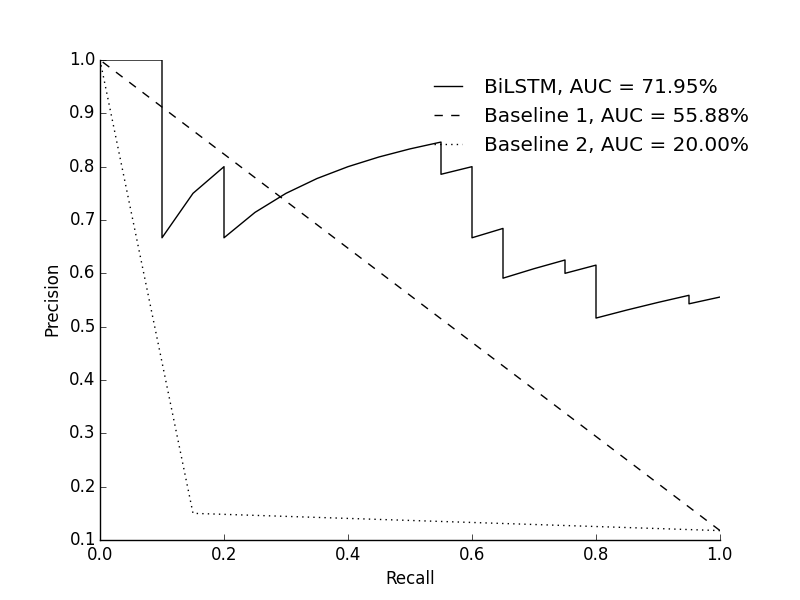} 
  \caption{The PR curve on the validation set using the optimal BiLSTM.}
  \label{fig: FINAL}
\end{figure}

The optimal hyperparameters are $10$ for the number of hidden nodes = $10$, and $0.01$ for the learning rate. The PR curve on the test set is shown in Figure \ref{fig: FINAL} and the PR AUC is $ 71.95\%$. To compare, two baselines were also evaluated. Baseline $1$ keeps predicting the same class all the time, resulting in a PR AUC of $55.88\%$. Baseline $2$ predicts the two classes according to the class imbalance, achieving a PR AUC of $20.00\%$. Both baselines perform significantly worse than the BiLSTM network.

\section{Conclusion}
This initial study investigated whether it is possible to use temporal information for predicting blood  culture test outcomes. A BiLSTM network was built taking as input a time sequence containing information from $3$ days and with a sampling frequency of one sample per hour. The output was a single number representing if there was a positive blood culture. Looking at the result, we can conclude that using temporal effects is useful in this setting.

Future work includes improving the network topology and comparing different types of networks that are able capture temporal effects, such as (bidirectional) recurrent neural networks or gated recurrent units. A direct comparison with non-temporal methods is necessary to truly examine the advantages of exploiting the temporal information in this data.

Other open problems include investigating the influence of the chosen hyperparameters such as the sample length and frequency, used to generate the time sequences. Especially interesting is the choice of the sampling end time. In this research, we defined it as the time when the first positive blood culture was taken, or as the last available point. However, one can choose the sampling end time to be an arbitrary time before the first positive samples are present. This would generate a clear benefit in a practical setting, as the system would be able to act as a decision support system and early detection algorithm, proposing the doctor to perform a test.

\section*{References}

\small

[1] Morrell, M., Fraser, V. J., \& Kollef, M. H.\ (2005). De-
laying the empiric treatment of candida bloodstream infection until positive blood culture results are obtained: a potential risk factor for hospital mortality. In
{\it Antimicrobial agents and chemotherapy 49}, pp.\ 3640--3645

[2] Ho, J. C., Lee, C. H. \& Ghosh, J.\ (2012) Imputation-enhanced prediction of septic shock in ICU patients. In {\it Proceedings of the ACM SIGKDD Workshop on Health Informatics}

[3] Mani, S., Ozdas, A., Aliferis, C., Varol, H.A., Chen, Q., Carnevale, R., Chen, Y., Romano-Keeler, J., Nian, H. and Weitkamp, J.H.,\ (2014). Medical decision support using machine learning for early detection of late-onset neonatal sepsis. In {\it Journal of the American Medical Informatics Association 21(2)}, pp.\ 326--336.

[4] Kim, J., Blum, J. M., \& Scott, C. D.\ (2010). Temporal features and kernel methods for predicting sepsis in postoperative patients. 

[5] Henry, K. E., Hager, D. N., Pronovost, P. J., \& Saria, S.\ (2015). A targeted real-time early warning score (TREWScore) for septic shock. In {\it Science Translational Medicine 7(299)},  pp.\ 1--9.

[6] Rangel-Frausto, M. S., Pittet, D., Costigan, M., Hwang, T., Davis, C. S., \& Wenzel, R. P.\ (1995). The natural history of the systemic inflammatory response syndrome (SIRS): a prospective study. In {\it Jama 273(2)}, pp.\ 117--123.

[7] Lipton, Z. C., Kale, D. C., \& Wetzel, R.\ (2016). Directly Modeling Missing Data in Sequences with RNNs: Improved Classification of Clinical Time Series. In {\it Machine Learning for Healthcare}, pp.\ 1–17

[8] Hochreiter, S., \& Schmidhuber, J. (1997).\ Long short-term memory. In {\it Neural computation 9(8)}, pp.\ 1735--1780.

\end{document}